\documentclass[runningheads]{llncs}
\usepackage{graphicx}
\usepackage{algorithmic}
\usepackage{algorithm}
\usepackage{multirow}
\usepackage{booktabs} 
\usepackage {amssymb}
\usepackage[colorlinks,linkcolor=blue]{hyperref}
\begin{document}
\title{CTS: A Consistency-Based Medical Image Segmentation Model}
%
%
\author{Kejia Zhang\inst{1} \and
Lan Zhang* \inst{1} \and Haiwei Pan \inst{1} \and Baolong Yu \inst{2}}
\authorrunning{Kejia Zhang et al.}
%
\institute{Harbin Engineering University \and
The Second Affiliated Hospital of Mudanjiang Medical University\\
\email{zhanglan2015@hrbeu.edu.cn}\\
}
\maketitle              
\begin{abstract}
In medical image segmentation tasks, diffusion models have shown significant potential. However, mainstream diffusion models suffer from drawbacks such as multiple sampling times and slow prediction results. Recently, consistency models, as a standalone generative network, have resolved this issue. Compared to diffusion models, consistency models can reduce the sampling times to once, not only achieving similar generative effects but also significantly speeding up training and prediction. However, they are not suitable for image segmentation tasks, and their application in the medical imaging field has not yet been explored. Therefore, this paper applies the consistency model to medical image segmentation tasks, designing multi-scale feature signal supervision modes and loss function guidance to achieve model convergence. Experiments have verified that the CTS model can obtain better medical image segmentation results with a single sampling during the test phase. 

\end{abstract}

\begin{figure}
\includegraphics[width=\textwidth]{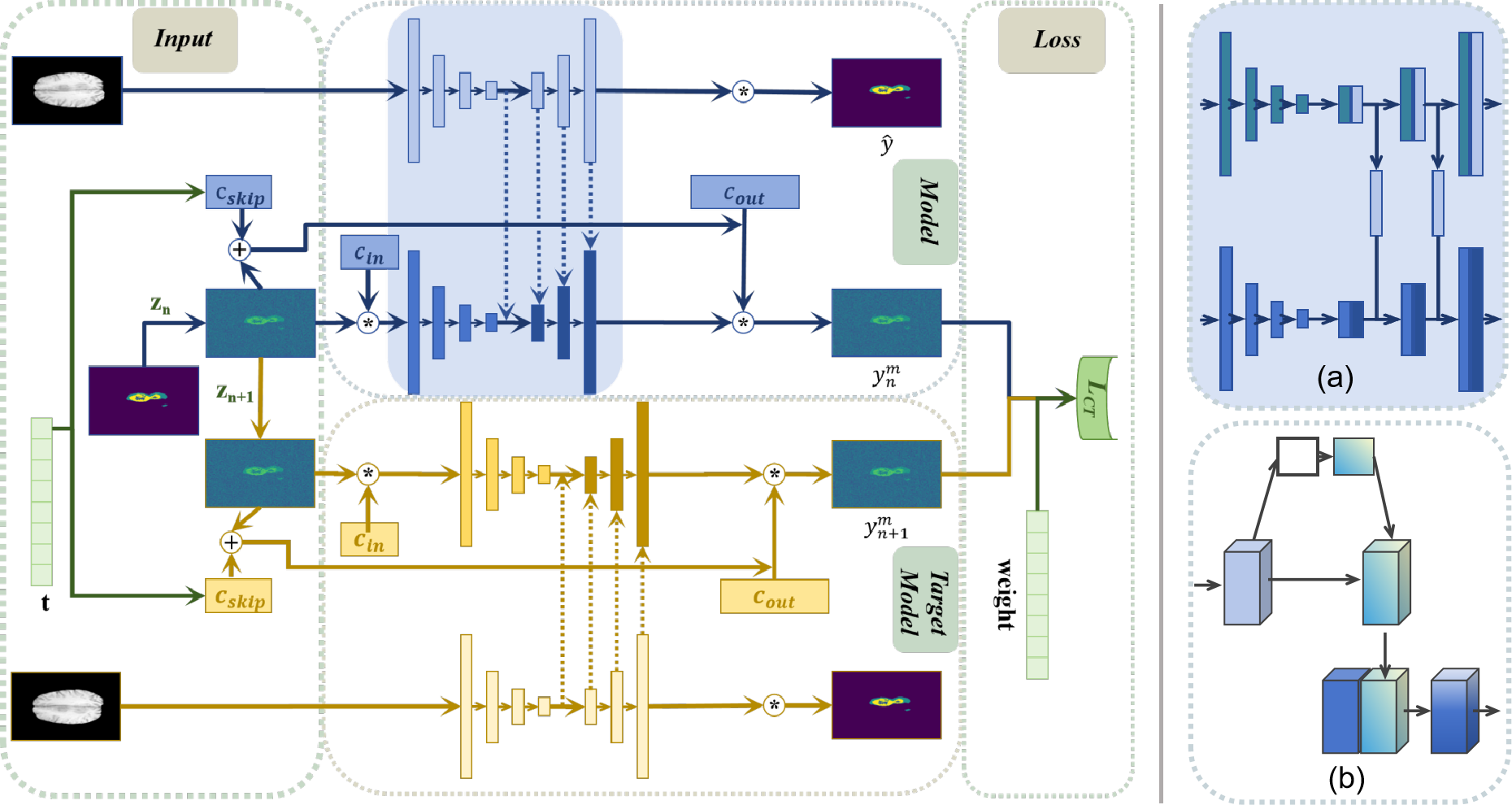}
\caption{CTS model overall flowchart. (a) The process of multi-scale feature supervision signal input is displayed. (b). The overlay process of feature supervision signals through channel attention mechanism is shown.} \label{whole}
\end{figure}
\section{Introduction}
The field of medical image segmentation has always been a hot research direction within the image segmentation domain. Unlike traditional segmentation methods\cite{2015U,unetplus}, utilizing generative models for image segmentation\cite{2017SegAN} can also achieve good results. Since diffusion models\cite{2015Deep,ho2020denoising} are a type of generative model that samples from Gaussian noise, the images they generate possess strong noise resistance and smoothness. Consequently, an increasing number of studies are leveraging diffusion models to tackle the non-generative issues of different images. Researchers use masks as the target for generative model sampling, while also incorporating constraints in the generative models to guide the direction of model generation. However, due to the need for extensive resampling during training and prediction, the issue of low computational efficiency in diffusion models urgently needs to be addressed.
Consistency model\cite{song2023consistency} transform multiple samplings into a single sampling by constructing a unique solution by ODE, significantly reducing the time consumed during the sampling process. Moreover, while reducing the number of samplings, consistency models also ensure the effectiveness of the samples. Compared to DDPMs\cite{ho2020denoising}, consistency model represent a superior generative paradigm, yet studies applying this model to the field of medical image segmentation are currently lacking. Therefore, this paper proposes constructing a medical image segmentation model based on the consistency model, and designing a loss function according to the segmentation loss and consistency training loss, enabling end-to-end training of the model. \textbf{CTS} code can be obtained in \href{https://github.com/LanHEU/CTS}{https://github.com/LanHEU/CTS} .
 
The specific contributions of this text are as follows:

\begin{itemize}
    \item A medical image segmentation model based on a consistency model has been constructed, featuring a newly designed joint loss function.
    \item During the decoding phase, multi-scale feature supervision signals are utilized to guide the model's convergence direction.
    
\end{itemize}

\section{Related Works}
In this section, we briefly describe the existing lines of research relevant to our work.
Diffusion models have been applied to many fields, such as sequence modeling \cite{li2022diffusion,chen2022analog}, speech processing\cite{popov2021grad}, computer vision\cite{saharia2022palette,ho2022imagen} to computed tomography (CT) scanning and magnetic resonance imaging (MRI). In computer vision, to reduce the number of sampling times, many methods have made great efforts. There are also some sampling algorithms tailored for conditional generation, such as without classifier guidance\cite{ho2022classifier} or with classifier guidance\cite{dhariwal2021diffusion}. Image segmentation is an important task in computer vision, which studies simplifying the complexity of an image by decomposing it into multiple meaningful image segments\cite{azad2022contextual,heidari2023hiformer}. Due to the time, cost, and expertise required\cite{azad2022transdeeplab,cao2022swin}, the number of images and labels for medical image segmentation is limited. For this reason, diffusion models, by synthesizing labeled data and eliminating the need for pixel-level labeled data, have become a promising method in image segmentation research. BrainSPADE\cite{fernandez2022can} proposed a generative model for synthesizing labeled brain MRI images, which can be used to train segmentation models. However, diffusion models in medical image segmentation face issues such as a high number of sampling times and long prediction times.

\section{Method}

This paper aims to fully leverage the advantages of sampling once with a consistency model, while retaining the benefits of the segmentation model. In consistency model\cite{song2023consistency}, the method of directly training a consistency model is referred to as consistency training loss, which is the origin of the 'CT' in the name of 'CTS'. The specific process is shown in Fig\ref{whole}.

Similar to the consistency model, the basic framework of this paper includes two parts: model $M$ and target model $TM$. The model's sampling begins with the mask $x^{m}$ of each image, inputting the corresponding data $x^{d}$as a supervisory signal.
Initialize the parameters of the two models, and copy the parameters from $M$ to $TM$. 

\textbf{Step 1}. The input to the model is the mask $x^{m}$, and noise $z_n$, sampled from a Gaussian distribution at step n, is added to the tensor: $x^{m}_n = x^{m} + z_n$.

\textbf{Step 2}. Simultaneously, based on the time period $t$, obtain: Model average learning moving parameters: $c_{in}$,$c_{out}$,$c_{skip}$

\textbf{Step 3}. The final input to model $M$ is: $x^{m}_{in} = c_{in} * x^{m}_n$

\textbf{Step 4}. Generate multi-scale signals using $\bigcup x_i^d,\hat{y}=h^T\left(x^d\right)$, where $i\in\left(1,2,...\right)$

\textbf{Step 5}. Here, the feature signal $\bigcup x_i^d$ is incorporated into the UNet model: $y_{out}^m=\ g^T\left(x_{in}^m,\bigcup x_i^d,t\right)$

\textbf{Step 6}. The output of the final model M is: $y_n^m=c_{out}\ast y_{out}^m+\ c_{skip}\ast x_n^m$

\textbf{Step 7}. Utilize the normal sampling method to obtain the noise$z_{n+1}$ of $\left(n+1\right)$th sampling from the Gaussian distribution. $x_{n+1}^m=f_\mathcal{U}\left(x^m,x_n^m,n,n+1\right)$

\textbf{Step 8}: Use $x_{n+1}^m$ as the input for model $TM$, and repeat the above \textbf{Step 2-5}. And the output is $y_{n+1}^m=c_{out}\ g^{TM}\left(x_{in}^m,x^d,t\right)\ +\ c_{skip}\ast x_{n+1}^m$

The consistency training segmentation loss is as follows: $\mathcal{L}_{CT}=yn+1m-ynm2$
To expedite the convergence speed and training outcomes, training is conducted on structures generated by multi-scale signals: $\mathcal{L}_S=y-xm2$
Overall Loss Function: $\mathcal{L}_{CTS}=\mathcal{L}_{CT}+\alpha\mathcal{L}_S$, where $\alpha$ is hyperparameter.

\textbf{Step 9}: Update the $TM$ model parameters; the $TM$ model update adheres to the learning rate: $\theta^{TM}\gets\ stopgrad\left(\mu\left(k\right)\theta^{TM}+\left(1-\mu\left(k\right)\right)\theta^M\right)$.

The pseudocode of the CTS algorithm is shown in Alg\ref{alg1}.

\textbf{Multi-scale Feature Supervision Signal}. The process of integrating multi-scale feature supervision signals $\bigcup x_i^d$ is shown in Fig\ref{whole}(a). The decoder stage of the image data encoding network progressively generates feature maps of each size, and combines them with the corresponding supervision signals.
In the decoder stage of the image data encoding network, a corresponding supervision signal $x_i^d$ is gradually generated for each size feature map. The supervision signal $x_i^d$, as shown in Fig\ref{whole}(b). This process is integrated into the $M$ model through a channel attention mechanism, achieving the addition of multi-scale supervision signals. During the decoder stage of the image data encoding network, these feature maps contain information of various scales, which can assist the model in better understanding the details and contextual information of the image. To better integrate the supervision signals and feature maps, a channel attention mechanism is employed. This automatically learns the importance weights of each channel, thereby making better use of the information from the supervision signals.

\begin{algorithm}[H]
\caption{Consistency Training Segmentation(CST)}
\begin{algorithmic}
\STATE 
\STATE \textbf{Input:} Dataset $\mathcal{D} $, initial model parameter $\theta^M $, learning rate $\eta$ , step schedule$ N ( \cdot ) $. EMA decay rate schedule $\mu (\cdot)$, $d(\cdot,\cdot)$,and $\lambda (\cdot)$
\STATE $\theta ^{TM}\leftarrow  \theta^M$ and $k\leftarrow 0$
\STATE \textbf{repeat}
\STATE \hspace{0.5cm} Sample $x^m,x^d \sim \mathcal{D} $, and $n \sim \mathcal{U} \left(1,N(k)-1\right) $
\STATE \hspace{0.5cm} Sample $z \sim \mathcal{N}\left(0,\mathbf{I} \right) $
\STATE \hspace{0.5cm} $\mathcal{L} _{CT}\left(\theta^M,\theta^{TM}\right) \leftarrow$
\STATE \hspace{0.6cm} $ \lambda(t_{n})d\{g_{\theta}^M[x+t_{n+1}\mathbf{z},h_{\theta}^M\left(x^d\right),t_{n+1} ],g_{\theta}^{TM}[x+t_{n}\mathbf{z},h_{\theta}^{TM}\left(x^d\right),t_{n} ]\}$
\STATE \hspace{0.5cm} $\theta \leftarrow - \eta\bigtriangledown_{\theta}\mathcal{L} _{CT}(\theta^M,\theta^{TM})$
\STATE \hspace{0.5cm} $\theta^- \leftarrow stopgrad(\mu(k)\theta^- + (1-\mu(k))\theta)$
\STATE \hspace{0.5cm} $k\leftarrow k+1$
\STATE \textbf{until} convergence
\end{algorithmic}
\label{alg1}
\end{algorithm}
\section{Experiment}
This section demonstrates through experimentation the advantages of CTS in medical image segmentation. We started with a thorough comparison of existing alternatives, followed by additional analysis to dissect the reasons behind CTS's success.

\textbf{Datasets}. We conducted experiments on medical tasks in two different image modalities: MRI image segmentation of brain tumors, and ultrasound image segmentation of thyroid nodules and liver tumor segmentation on the BraTs-2021 dataset\cite{baid2021rsna} as well as in SEHPI datasets\cite{zhang2023sunet++}. This paper utilized anisotropic diffusion filtering\cite{passalacqua2010geometric}, while also removing Poisson noise from medical images, preserving more edge information and effective feature structures. Consequently, this further improved the model's performance.

\begin{table*}
\centering
\caption{Result. CTS-nM: without multiscale feature supervision signals. CTM-M: with multiscale feature supervision signals. CTM-FM: with the FFTP structure mentioned same with MedSegDiff.} \label{result1}
\begin{tabular}{c|cc|cc|cc}
 \toprule
\multirow{2}{*}{} & \multicolumn{2}{c|}{Brain-Turmor} & \multicolumn{2}{c|}{SEHPI}    & \multicolumn{2}{c}{Thyroid Nodule} \\
                  & Dice            & IoU             & Dice          & IoU           & Dice             & IoU             \\ \midrule
CENet             & 76.2            & 68.9            & 82.4          & 70.5          & 78.9             & 71.2            \\
MRNet             & 83.4            & 75.6            & 85.9          & 73.4          & 80.4             & 73.4            \\
SegNet            & 80.2            & 72.9            & 86.8          & 73.1          & 81.7             & 74.5            \\
nnUNet            & 88.2            & 80.4            & 88.2          & 72.9          & 84.2             & 76.2            \\
TransUNet         & 86.6            & 79.0            & 89.3          & 75.4          & 83.5             & 75.1            \\
MedSegDiff-L      & 89.9            & 82.3            & 89.8          & 78.5          & 86.1             & 79.6            \\ \hline
CTS-nM            & 90.0            & 82.5            &  -            &     -         &   -               &  -              \\
MedSegDiff++      & 90.5            & 82.8            & 90.3          & 79.3          & 86.6             & 80.2            \\
CTS-M             & 91.7            & 83.9            & \textbf{91.0} & \textbf{80.5} & \textbf{87.3}    & \textbf{81.3}   \\
CTS-FM            & \textbf{92.1}   & \textbf{84.0}   &  -            &     -         &   -               &  -             \\
\bottomrule
\end{tabular}
\end{table*}

\textbf{Experiment Details}. We utilized a $4\times$ UNet. In the testing phase, we employed a single diffusion step for inference, which is significantly smaller than most previous studies. All experiments were implemented using the PyTorch platform and executed on one GTX 4090. All images were uniformly resized to 256×256 pixels. Training was conducted in an end-to-end manner using the AdamW optimizer. $batchsize=8$. The learning rate was initially set to $1 \times 10^{-4}$.
\begin{figure}
\includegraphics[width=\textwidth]{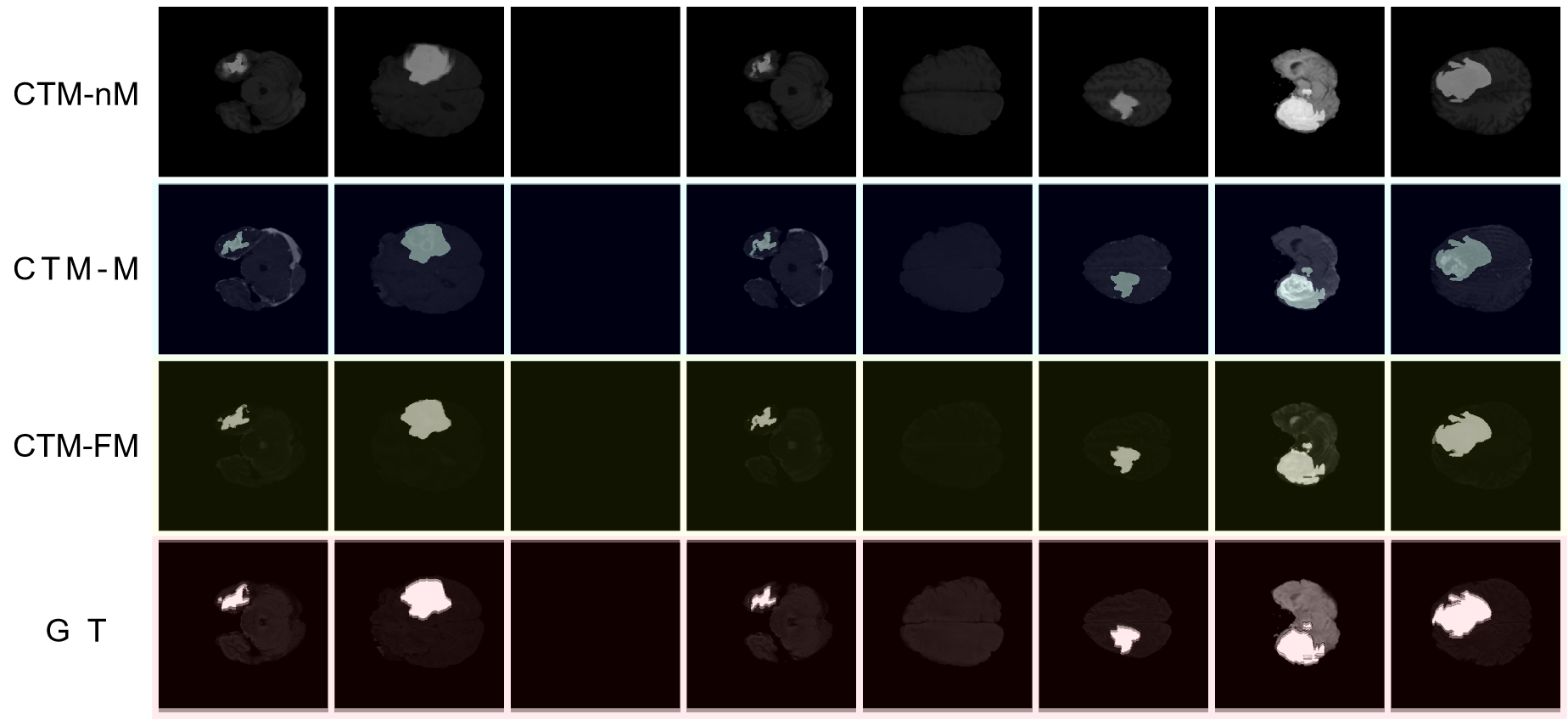}
\caption{Result Visualization} \label{vison}
\end{figure}

\textbf{Main result}. We compared the SOTA segmentation methods proposed for each task with general medical image segmentation methods. The main results are presented in Table 1. Part of the relevant results originates from the work\cite{wu2024medsegdiff}.
In our experiments, we trained on each dataset for 700,000 iterations, with the specific training duration being one month.
CTS-nM indicates that the model did not use multiscale feature supervision signals. CTM-M denotes the use of multiscale feature supervision signals. For a fair comparison, the Meg method incorporates a Fourier filter, while CTM-FM includes the FFTP structure mentioned same with MedSegDiff. Detailed results can be seen in Tab\ref{result1}. Visualization results are shown in Fig\ref{vison}.
The results reveal that CTS-nM can surpass most methods, demonstrating that consistency model can not only reduce the number of samplings but also enhance effectiveness. The performance of the CTM-M model, with the addition of multi-scale feature signals, is further improved. CTM-FM also proves that Fourier filtering can further enhance performance. All methods under CTS were tested with a single sampling, averaging 1.9s. This significantly reduces the time compared to other models. Therefore, the CTS model can guarantee model effectiveness while accelerating sampling.

\begin{figure}
\begin{minipage}[t]{0.48\linewidth}
\centering
\includegraphics[width=\textwidth]{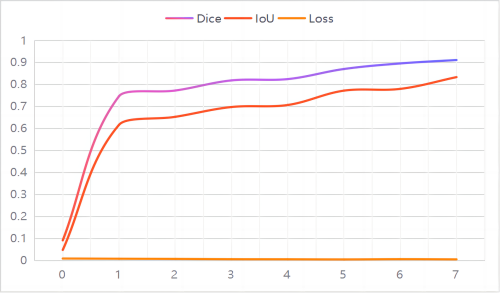}
\caption{Models saved at different stages, their training loss, and corresponding results on the testset.} \label{dice}
\end{minipage}
	\quad
\begin{minipage}[t]{0.48\linewidth}
\centering
\includegraphics[width=\textwidth]{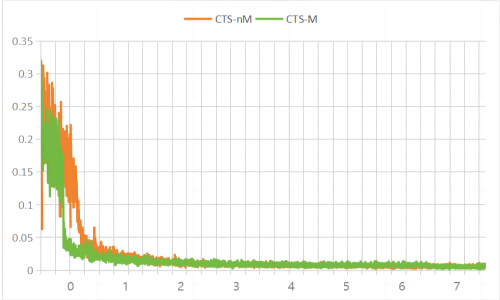}
\caption{Accelerating the convergence speed of multi-scale feature signal models} \label{loss}
\end{minipage}
\end{figure}

In Fig\ref{dice}, the convergence process of the model is shown. The trend of the loss value during training. The changing trends of IoU and Dice metrics for model parameters saved at different times on the test set. It can be observed that, as the training time increases, the loss region during training becomes more leveled, but the test results do not show saturation, with a significant growth margin. This is likely related to the strong learning and representational capabilities of the consistency model. The findings of this paper are in agreement with the conclusions of the consistency studies.


\begin{table*}[]
\centering
\caption{Ablation experiment. Mul\_s represents the multi-scale supervisory signal. FFTP denotes the Fourier filter structure.} \label{result2}
\begin{tabular}{cc|cc|cc}
\toprule
\multicolumn{2}{c|}{} & \multicolumn{2}{c|}{Brain-Turmor} & \multicolumn{2}{c}{SEHPI} \\
Mul\_s      & FFTP       & Dice            & IoU             & Dice        & IoU         \\ \midrule
           &            & 85.3            & 79.2            & 78.2        & 85.1        \\
\checkmark &            & 86.3            & 80.1            & 79.1        & 84.9        \\
           & \checkmark & 86.2            & 80.3            & 79.3        & 85.6        \\
\checkmark & \checkmark &\textbf{86.9}    &\textbf{82.1}    &\textbf{81.1}&\textbf{86.3} \\ \bottomrule
\end{tabular}
\end{table*}
\textbf{Ablation experiment} Fig\ref{loss} compares the convergence speeds between CTS-nM and CTS-M. Since models typically enter a smoothing phase after exceeding 10,000 rounds, only the experimental results of 20,000 rounds are compared here. It is evident that the inclusion of multi-scale feature signals significantly accelerates the convergence speed of the models.Tab\ref{result2} compares the ablation results between two datasets. Here, to observe the result, each experiment was trained with 500,000 samples. It is evident that incorporating multi-scale signals yields better results. Additionally, FFTP can also enhance performance. Adding either element individually does not significantly differ in improving the model, possibly because both involve modifications to the way supervisory signals are added, with the difference lying in the method. This further indicates that adding better supervisory signals can greatly impact the model's outcomes. Eventually, the model that incorporated both methods was used to train the overall model.

\section{Conclusion and Discussion}
This paper first establishes a medical image segmentation method based on a consistency model \textbf{CTS}. It not only yields better results but also significantly reduces prediction time. Moreover, by constructing multi-scale feature supervision signals, the training convergence speed is accelerated. Meanwhile, medical image data processed with anisotropic edge enhancement filters can achieve improved outcomes. However, there are still some shortcomings. In our experiments, due to a lack of GTX 4090, we did not train as many as one million rounds as initially planned in the Consistency\cite{song2023consistency}. Furthermore, we observed that increasing the number of training rounds did not lead to saturation in results, leading us to infer that more training rounds could potentially enhance the model's performance.

\bibliographystyle{splncs04}
\bibliography{name}
%




\end{document}